%% file: main.tex
\pdfoutput=1
\documentclass[10pt,twocolumn,letterpaper]{article}

\usepackage[pagenumbers]{cvpr}

\definecolor{cvprblue}{rgb}{0.21,0.49,0.74}
\usepackage[pagebackref,breaklinks,colorlinks,allcolors=cvprblue]{hyperref}

\usepackage{multirow}
\usepackage{xcolor}
\usepackage{colortbl}
\usepackage{pdfrender}

\newcommand{\best}[1]{%
  \textcolor{red}{\textpdfrender{TextRenderingMode=FillStroke,LineWidth=0.15pt}{\textbf{#1}}%
  }%
}
\newcommand{\second}[1]{\textcolor{blue}{\underline{#1}}}

\title{TwinScape-GS: Virtual Flyovers for City-Scale Surface Reconstruction}

\author{Tengkai Wang \qquad Daochang Liu \qquad Saeed Anwar \qquad Ajmal Saeed Mian\\
The University of Western Australia
}

\begin{document}

\twocolumn[{%
  \renewcommand\twocolumn[1][]{#1}%
  \maketitle
  \centering
  \includegraphics[width=\linewidth]{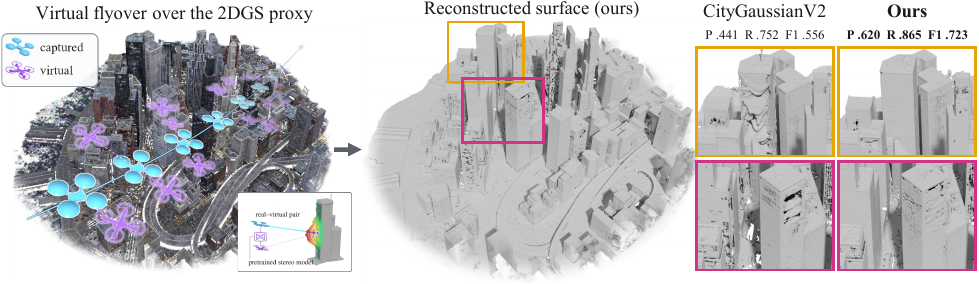}
  \captionsetup{hypcap=false}
  \vspace{-5mm}
  \captionof{figure}{TwinScape-GS reconstructs accurate city-scale surfaces through virtual flyovers. From the captured views (blue drones), we build a 2DGS twin of the city as a geometry proxy, then plan a virtual flyover (purple drones) that revisits poorly reconstructed surfaces. Each planned virtual camera is paired with a captured view, allowing a pretrained stereo model to estimate depth that complements the geometry reconstructed from the captured views. Compared with CityGaussianV2, our reconstruction produces flatter building facades and fewer geometric artifacts, with much higher precision (P), recall (R), and F1 score.}
  \label{fig:teaser}
  \vspace{1em}
}]

\input{sec/0_abstract}
\input{sec/1_introduction}
\input{sec/2_related_work}
\input{sec/3_preliminaries}
\input{sec/4_method}
\input{sec/5_experiments}
\input{sec/6_conclusion}

\section*{Acknowledgement}
This research is supported by the Australian Research Council (ARC) under the Discovery Project scheme (Project ID: DP240101926).

{
    \small
    \bibliographystyle{ieeenat_fullname}
    \bibliography{main}
}

\input{sec/7_supplementary}

\end{document}

%% file: sec/0_abstract.tex
\begin{abstract}
Recent Gaussian Splatting-based methods have achieved high-fidelity novel view synthesis in city-scale scenes. However, recovering geometrically complete and accurate surfaces remains challenging because aerial captures are locally sparse despite being globally dense. Planned trajectories usually allow aerial capture of many local surfaces from only a limited range of directions, with insufficient constraints for reliable triangulation. Existing city-scale methods derive geometry from the captured views only; hence, their accuracy is limited by the original viewing configuration. We present TwinScape-GS, a 2D Gaussian Splatting (2DGS)-based method that reconstructs high-quality city-scale surfaces through virtual flyovers. By planning virtual cameras, TwinScape-GS combines the novel view synthesis of Gaussian Splatting with the strong prior of a stereo model to recover poorly constrained surfaces from new viewing configurations. We first design a disparity-based stereo pairing scheme to build a 2DGS city proxy from the capture alone. Over this proxy, we propose a multi-view consistency-guided virtual flyover planner that allocates virtual stereo views to the surfaces that are poorly reconstructed. Experiments on large-scale urban datasets GauU-Scene and MatrixCity demonstrate that TwinScape-GS achieves superior reconstruction performance compared to state-of-the-art approaches. We will release code, weights, and results.
\end{abstract}

%% file: sec/1_introduction.tex
\vspace{-2mm}
\section{Introduction}
\vspace{-1mm}
\label{sec:intro}

3D scene reconstruction is a fundamental computer vision task. At city scale, reconstructing scenes with \textit{accurate geometry} is the key to a wide range of applications, including aerial surveying~\cite{urbanscene3d,gauuscene}, urban planning~\cite{citymodels_review}, and autonomous navigation~\cite{kitti,nuscenes}. Recently, 3DGS~\cite{3dgs} has reshaped the task, owing to its photorealistic rendering and efficient training. However, appearance can be well modeled by fitting Gaussian primitives even at incorrect positions.
On the other hand, accurate geometry needs sufficient triangulation constraints.

In urban environments, geometry reconstruction faces a paradoxical challenge: \textit{the captured data are globally dense but locally sparse}. Ideally, every surface should be observed from sufficiently diverse directions, but across a complex, city-scale layout, capture cost and limited time windows put such acquisition out of reach. In practice, the planned trajectories generally cover the global structure densely, but local surfaces may still be observed from only a limited range of directions, leaving insufficient constraints for reliable triangulation.

To improve surface reconstruction at the city scale, existing methods typically extend 2DGS~\cite{2dgs} and PGSR~\cite{pgsr} to large-scale scenes with additional regularization. Some incorporate monocular depth priors for dense guidance, but these predictions are scale-ambiguous and estimated independently for each view. Others return to classical multi-view geometry, using geometric consistency checks~\cite{citygsx}, iterative depth refinement such as PatchMatch~\cite{metrogs,comvsgs}, or view grouping~\cite{vopgs} to improve surface accuracy. However, both strategies optimize geometry using only the captured views rather than introducing new viewpoints. As a result, their improvements are limited by the original viewing configuration.

In contrast, model-based view planning methods plan new real captures based on a coarse scene proxy built from an initial flight, such as a mesh~\cite{smith2018,learning_reconstructability},  and more recently target radiance field reconstruction~\cite{openflyscan}. These methods use the proxy geometry to place new viewpoints that maximize reconstruction accuracy, but still require a second physical flight. A few large-scale methods instead add virtual views to complement the capture, but typically place them along fixed trajectories and then repair them with generative diffusion priors~\cite{lee2025skyfall,zhang2026dpgs,yu2026orbit}. They generally focus on generative appearance modeling at the cost of hallucinated geometry. This raises a natural question: can we virtually fly over the reconstruction itself to recover accurate geometry?

Stereo depth priors estimate depth from correspondences between two views, without scale ambiguity. They have recently proven effective in small-scale or object-centric scenes~\cite{gs2mesh,safadoust2024self,stereogs,eve3d}, where methods typically form a stereo pair with rendered views at a heuristic offset from each captured view. However, this placement strategy does not adapt to the local scene geometry and thus does not scale to city-scale scenes, which are vast and geometrically complex, with substantial occlusion and significant height variation.

Based on these observations, we present TwinScape-GS, which reconstructs accurate city-scale surfaces through virtual flyovers (Fig.~\ref{fig:teaser}). Instead of physically flying the scene again, we plan virtual cameras over its twin, a 2DGS~\cite{2dgs} proxy of the captured scene, render them by novel view synthesis, and pair each with a captured view to recover geometrically accurate depth using a pretrained stereo model~\cite{foundationstereo}. Specifically, we first design a disparity-based stereo pairing scheme to obtain accurate stereo depth from the captured views. It pairs the captured views to maximize disparity of both nearby and distant surfaces within the matching range of the stereo model. The resulting stereo depth trains the 2DGS proxy. Over this proxy, we propose a multi-view consistency-guided virtual flyover planner. It locates poorly reconstructed surfaces from the multi-view consistency of rendered depths and allocates a limited number of virtual cameras to them. Finally, we jointly supervise the 2DGS model with the stereo depth from both types of pairs. In summary, our contributions are as follows:
\begin{itemize}
    \item We present TwinScape-GS, which reconstructs accurate city-scale surfaces through virtual flyovers. To our knowledge, it is the first radiance field-based method to plan virtual views for city-scale surface reconstruction.
    \item We design a disparity-based stereo pairing scheme that builds a 2DGS proxy from the captured views. On this proxy, we propose a multi-view consistency-guided virtual flyover planner that allocates a limited number of virtual cameras to poorly reconstructed surfaces.
    \item Experiments on GauU-Scene~\cite{gauuscene} and MatrixCity~\cite{matrixcity} show that TwinScape-GS reconstructs higher-quality surfaces than state-of-the-art methods.
\end{itemize}

%% file: sec/2_related_work.tex
\section{Related Works}
\label{sec:related}

\noindent {\bf Surface Reconstruction.} 3D Gaussian Splatting (3DGS)~\cite{3dgs} enables real-time rendering with explicit Gaussian primitives, but its unstructured, volumetric primitives have no well-defined surface, which leads to ambiguous depth and noisy geometry~\cite{2dgs,sugar}. Recently, many methods have adapted 3D Gaussians to surface reconstruction. Some replace volumetric primitives with surface primitives, such as planar surfels~\cite{2dgs,gaussiansurfels} or quadric surfaces~\cite{qgs}. Others keep 3D Gaussians and define a surface over them to estimate depth or generate a mesh~\cite{sugar,gof,radegs,gggs,milo}. Many studies also incorporate geometric constraints that cannot be provided by photometric loss alone, including normal consistency~\cite{2dgs}, monocular depth and normal priors~\cite{dnsplatter}, and multi-view consistency~\cite{pgsr,gaussurf}. These methods mainly improve geometric accuracy on object-level and small-scale scenes. Among them, 2DGS provides view-consistent depth through ray--splat intersection and is widely used as the basis of subsequent work, including city-scale methods~\cite{citygaussianv2,metrogs}. Our method also builds on 2DGS.

\vspace{1mm}\noindent \textbf{Large-Scale Scene Reconstruction.} Large-scale Gaussian Splatting methods improve rendering quality and efficiency through scene partitioning~\cite{vastgaussian,citygaussian}, hierarchical representations~\cite{hierarchicalgs,octreegs,alodofgaussians}, and distributed training~\cite{dogaussian,grendel}. To reconstruct surfaces at city scale, recent methods extend 2DGS~\cite{2dgs} and PGSR~\cite{pgsr} to large scenes with partitioned or distributed training~\cite{citygaussianv2,cosurfgs,citygsx,gigags,vopgs}. Some add monocular depth or normal priors~\cite{citygaussianv2,urbangs,metrogs}, which are scale-ambiguous and multi-view inconsistent. Others use multi-view constraints across captured views, such as multi-view consistency~\cite{gigags,ulsrgs,streetsurfgs,citygsx} or PatchMatch depth refinement~\cite{metrogs,comvsgs}. All of these methods constrain geometry using only the captured views.

\vspace{1mm}\noindent\textbf{View Planning and Virtual Views with Pretrained Priors.}
View planning selects viewpoints from which a scene can be reconstructed accurately. Model-based methods~\cite{roberts2017submodular,smith2018,learning_reconstructability,openflyscan} build a coarse proxy of the scene from an initial flight and plan viewpoints for a second flight on it. They maximize surface coverage~\cite{roberts2017submodular} or a reconstructability score computed from the viewing geometry at each surface point~\cite{smith2018,learning_reconstructability}. Some methods recapture poorly reconstructed regions~\cite{mostegel,onthefly_sfm} or select next views by radiance-field uncertainty~\cite{activegs,fisherrf,auto3r}. All of these methods need physical captures at the planned viewpoints. Without new captures, other methods render virtual views from a reconstructed Gaussian model and supervise them with pretrained priors. Some use generative priors. For example, Difix3D+~\cite{difix3d} repairs artifacts in rendered views with a diffusion model and distills the repaired views back into the reconstruction. At large scale, recent methods~\cite{lee2025skyfall,zhang2026dpgs,yu2026orbit} generally render virtual views along fixed trajectories and repair them with diffusion models to complete regions that the capture does not observe. Others use stereo priors. GS2Mesh~\cite{gs2mesh} renders a stereo pair at each training view with a fixed offset proportional to the scene radius, and fuses the stereo depth into a mesh. Later methods use stereo depth to supervise training~\cite{stereogs,eve3d,safadoust2024self,gu2026sparsesurf}, and FDS~\cite{fds} distills a matching prior between captured and virtual views. These methods are evaluated on small-scale scenes and generally place the virtual view at a heuristic baseline from each captured view. In contrast, our method plans virtual cameras instead of physical captures, and places them according to the poorly reconstructed surfaces of a city-scale scene rather than heuristically.

%% file: sec/3_preliminaries.tex
\section{Preliminaries}

\begin{figure*}[t]
    \centering
    \includegraphics[width=\linewidth]{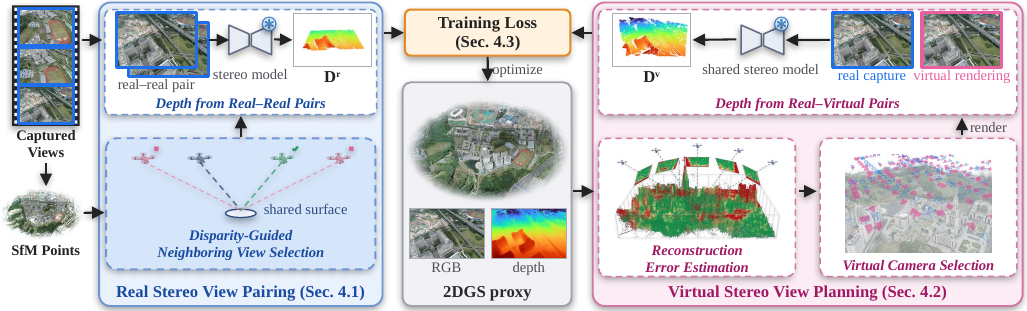}
    \caption{\textbf{Overview of TwinScape-GS.} Starting from the captured views, we select neighboring views that give their shared SfM points a large but still matchable disparity for accurate depth estimation, and train a 2DGS proxy with the depth $D^{r}$ that the stereo model estimates from the resulting real--real pairs (Sec.~\ref{subsec:real-real}). The virtual flyover is planned over this proxy. Our multi-view consistency-guided virtual flyover planner estimates the reconstruction error from the consistency of depths rendered from the proxy across neighboring views and back-projects it to 3D points to locate poorly reconstructed surfaces. These surfaces guide the virtual camera selection. During the flyover, the virtual rendering at each selected camera is paired with a real capture, and the shared stereo model estimates the depth $D^{v}$ of this real--virtual pair (Sec.~\ref{subsec:real-virtual}). Finally, $D^{v}$ from the flyover complements $D^{r}$ from the capture in the training loss by supervising each pixel with the depth of the highest multi-view consistency to optimize the proxy (Sec.~\ref{subsec:train_loss}).}
\label{fig:pipeline}
\end{figure*}

\textbf{3D Gaussian Splatting} \cite{3dgs} represents the scene as a set of Gaussian spheres. Each sphere is parameterized by position $\boldsymbol{\mu}$, covariance $\boldsymbol{\Sigma}$, opacity $o$, and view-dependent color $\mathbf{c}$.  When rendering, 3DGS effectively splats the color $C$ for a given pixel $p$ by alpha blending a set of Gaussian spheres:

\begin{equation}
\begin{aligned}
C(p) &= \sum_{i \in \gamma(p)} \mathbf{c}_i \alpha_i T_i, \qquad T_i = \prod_{j=1}^{i-1} (1-\alpha_j), \\
\alpha_i &= o_i \exp\!\Big(-\tfrac{1}{2}(p-\boldsymbol{\mu}'_i)^{\top}\boldsymbol{\Sigma}'^{-1}_i(p-\boldsymbol{\mu}'_i)\Big),
\end{aligned}
\end{equation}
where $\gamma(p)$ is the depth-sorted set of Gaussians overlapping $p$; $T_i$ is the transmittance in front of the $i$-th Gaussian, and $\boldsymbol{\mu}'_i$, $\boldsymbol{\Sigma}'_i$ are its mean and covariance projected onto the image plane.

\noindent \textbf{2D Gaussian Splatting}~\cite{2dgs} flattens the 3D Gaussian spheres into 2D disks. During training, it employs depth distortion regularization along with a normal consistency loss that aligns surfel normals with those derived from the rendered depth. This encourages a surface-aligned and depth-meaningful scene representation. Given an optimized 2DGS representation, the median depth is rendered as the z-depth of the last disk along the ray whose transmittance exceeds one half, thereby approximating the ray-surface intersection:
\begin{equation}
\tilde{D}(p) = \max\{\, z_i \mid T_i > 0.5 \,\},
\end{equation}
where $z_i$ is the depth of the $i$-th Gaussian along the ray.

\noindent \textbf{Stereo Matching.} A stereo pair consists of a reference view $I_i$ and a second view $I_j$ with the distance between their camera centers equal to the baseline $b$. The pair is first rectified so that corresponding pixels lie on the same row~\cite{hartley2004multiple}. A stereo model then estimates the disparity $d(p)$ for each pixel $p$ in $I_i$, defined as the horizontal offset to its corresponding pixel in $I_j$. The depth of $p$ is
\begin{equation}
    z(p) = \frac{f\,b}{d(p)},
    \label{eq:disparity}
\end{equation}
where $f$ is the focal length. A larger disparity gives a more accurate depth, since a disparity error $\epsilon_d$ leads to a relative depth error
\begin{equation}
    \frac{\epsilon_z}{z} \approx \frac{\epsilon_d}{d}.
    \label{eq:depth_error}
\end{equation}
A longer baseline increases the disparity and makes the two views harder to match~\cite{okutomi1993multiple}, because correspondences must be searched over a larger disparity range and more of the scene is occluded or appears differently in the two views.

%% file: sec/4_method.tex
\section{TwinScape-GS}
\label{sec:method}

TwinScape-GS builds on 2DGS and consists of two modules as illustrated in Fig.~\ref{fig:pipeline}. We pair captured views into real--real stereo pairs and train a 2DGS proxy of the city with their depth, estimated by a stereo model (Sec.~\ref{subsec:real-real}). We then plan a virtual flyover over this proxy that adds real--virtual stereo pairs for poorly reconstructed surfaces (Sec.~\ref{subsec:real-virtual}), and jointly supervise the GS model with the stereo depth from both types of pairs (Sec.~\ref{subsec:train_loss}).

\subsection{Real Stereo View Pairing}
\label{subsec:real-real}

\textbf{Motivation.} For each captured view $I_i$, we pair it with neighboring views so that the stereo model can estimate its depth as accurately as possible. A stereo pair gives a more accurate depth for a surface when its disparity is larger (Eq.~\ref{eq:depth_error}), as long as it remains small enough to match. Existing city-scale methods~\cite{metrogs,citygsx} typically select several nearby views with similar viewing directions as neighbors for patch matching. However, their short baselines produce small disparities, leading to larger depth errors. Simply increasing the baseline is not sufficient either. At city scale, a single captured image often covers a wide range of depths, from nearby rooftops to distant ground. A longer baseline improves depth accuracy for distant surfaces but makes nearby surfaces harder to match. Therefore, rather than selecting neighboring views based on camera-pose similarity, we select complementary neighboring views based on the disparities they produce, so that both nearby and distant surfaces in $I_i$ can be observed with large yet matchable disparities.

\vspace{1mm}\noindent \textbf{Neighboring View Selection.} Given the captured images $\{I_i\}_{i=1}^{N}$, we represent the surfaces observed in each view $I_i$ by its visible SfM~\cite{schoenberger2016sfm} points $\mathcal{P}_i$. We then collect all views that share SfM points with $I_i$ as its neighboring views $\mathcal{N}_i$.

For each neighboring view $j\in\mathcal{N}_i$, we rectify the pair
$(I_i, I_j)$ and compute the expected disparity $d_{ij}(\mathbf{X})$ at each shared SfM point $\mathbf{X}$ by projecting it into both rectified images. MVSNet~\cite{yao2018mvsnet} selects neighboring views whose baseline angle at shared SfM points is large enough for triangulation but small enough to match, around a manually set angle. We follow the same criterion on the disparity, but maximize it up to the limit given by the stereo model. We define the pair score of a disparity $d$ as:
\begin{equation}
    q(d) = \mathbf{1}\!\left[d_{\min} \le d \le d_{\max}\right] \cdot \frac{d}{d_{\max}},
    \label{eq:q}
\end{equation}
where $d_{\max}$ is the largest disparity the stereo model searches, and $d_{\min}$ is the smallest disparity we accept. Within this range, the score increases with the disparity. The score of view $j$ for point $\mathbf{X}$ is then $q_{ij}(\mathbf{X}) = q(d_{ij}(\mathbf{X}))$.

We choose the stereo pairs for $I_i$ by selecting a subset
$\mathcal{G}\subseteq\mathcal{N}_i$ of at most $K$ views:
\begin{equation}
    \mathcal{G}_i =
    \arg\max_{\substack{
        \mathcal{G}\subseteq\mathcal{N}_i\\
        |\mathcal{G}|\le K
    }}
    \sum_{\mathbf{X}\in\mathcal{P}_i}
    \max_{j\in\mathcal{G}} q_{ij}(\mathbf{X}),
    \label{eq:select}
\end{equation}
where each point counts only its best pair in $\mathcal{G}$, so that the selected views complement each other. We solve it by greedy selection.

\vspace{1mm}\noindent \textbf{Depth from Real--Real Pairs.} For each selected neighboring view $j \in \mathcal{G}_i$, we rectify the image pair $(I_i, I_j)$ as required by FoundationStereo~\cite{foundationstereo}, which then predicts the disparity map of $I_i$. We discard the pixels that fail the left--right consistency check or whose disparity is outside $[d_{\min}, d_{\max}]$, convert the remaining disparities to depth using Eq.~\ref{eq:disparity}, and fuse the resulting $|\mathcal{G}_i|$ depth maps into a single depth map $D_i^{r}$ by taking the per-pixel median over the valid depths. These depth maps supervise a 2DGS model of the scene, which serves as the proxy for the virtual flyover in Sec.~\ref{subsec:real-virtual}.

\subsection{Virtual Stereo View Planning}
\label{subsec:real-virtual}

\textbf{Motivation.} Existing city-scale methods~\cite{metrogs,citygaussianv2,citygsx} optimize scene geometry only from the captured camera configuration. Model-based view planning methods~\cite{smith2018} build a proxy from an initial capture and plan a second physical flight, whereas we plan a virtual flyover on the proxy. Specifically, we use the 2DGS model to plan virtual cameras and pair them with captured images. Because the second view of each pair is rendered, its baseline is no longer constrained to those available in the capture. The planner selects virtual cameras that produce a large, yet still matchable, disparity for poorly reconstructed surfaces.

\vspace{1mm}\noindent \textbf{Reconstruction Error Estimation.} A surface point subject to insufficient triangulation constraints has greater uncertainty in its reconstructed 3D position~\cite{hartley2004multiple,lim2023fisher}. In a 2DGS model, such uncertainty can appear as inconsistent depth estimates across rendered views. Without ground-truth geometry to measure surface reconstruction errors, we therefore propose using the multi-view consistency of rendered depths, which is widely used in multi-view stereo to evaluate depth estimates~\cite{schoenberger2016pixelwise,yao2018mvsnet},
to estimate how accurately each local surface has been reconstructed.

Given a depth hypothesis $Z$ in the reference view $I_i$ and a set of neighboring depth maps $\mathcal{Z}=\{Z_k\}_{k\in\mathcal{N}_i}$, we compute the forward--backward reprojection error $e_{ik}(p)$ for each pixel $p$. Specifically, we warp $p$ to $I_k$ using $Z(p)$, then warp it back to $I_i$ using $Z_k$. We then measure the multi-view consistency of $Z(p)$ by summing a Gaussian kernel of this error across the valid neighboring views:
\begin{equation}
Q_i(p; Z, \mathcal{Z}) =
\sum_{k \in \mathcal{N}_i(p)}
\exp\left(-\frac{e_{ik}(p)^2}{2\sigma^2}\right),
\label{eq:vote}
\end{equation}
in which $\mathcal{N}_i(p)$ is the set of neighboring views where $p$ has a valid depth and projection, and $\sigma$ controls how much reprojection error is tolerated. $Q_i$ is higher for depths that more views agree on.

For each captured view $I_i$, we render its depth map $\tilde{D}_i$ from the 2DGS scene proxy and compute its multi-view consistency $Q_i$ against the rendered depths of its neighboring views with Eq.~\ref{eq:vote}. We then convert $Q_i$ into a per-pixel estimated reconstruction error $E_i(p)=Q_{\max}-Q_i(p)$, where $Q_{\max}$ is the maximum of $Q_i(p)$ over all captured views. We back-project the pixels that have at least one valid neighboring view using the rendered depths to obtain a set of 3D surface points $\mathcal{P}_e$, where each point $\mathbf{X}$ inherits the estimated error of its pixel, $w(\mathbf{X})=E_i(p)$.

\vspace{1mm}\noindent \textbf{Virtual Camera Selection.} Following virtual stereo methods~\cite{stereogs,eve3d}, we construct each real--virtual pair with a real capture as the reference view and a virtual rendering as the second view, so that the reference view is free of rendering artifacts and optimization noise. Previous methods~\cite{stereogs,gu2026sparsesurf,gs2mesh} use only one virtual view for each captured view, at a single fixed baseline. However, as discussed in Sec.~\ref{subsec:real-real}, a baseline that is long enough for distant surfaces usually makes nearby surfaces harder to match. Therefore, we adopt an oversample-then-select strategy, sampling candidate virtual cameras at multiple baselines for each captured view and selecting among them according to the estimated reconstruction error.

Specifically, given a captured view, the corresponding virtual camera keeps its orientation and is translated by a baseline $b$ along the camera's $x$-axis. We adapt the baseline to the scene depth of each captured view as $b = \delta\,\bar{z}_i / f$, where $\bar{z}_i$ is its median rendered depth and $\delta$ is the resulting disparity at the median depth. We sample a set of $\delta$ values for each captured view. Each combination of a captured view and a value of $\delta$ gives a candidate virtual camera, and we denote the set of all candidates by $\mathcal{V}$.

We then score each candidate $v \in \mathcal{V}$ on the surface points $\mathcal{P}_e$ in the same way as in Sec.~\ref{subsec:real-real}. For a point $\mathbf{X}$ at depth $z$ in the captured view of $v$, the expected disparity is $d_v(\mathbf{X}) = f b_v / z$, where $b_v$ is the baseline of $v$, and its score is $q_v(\mathbf{X}) = q(d_v(\mathbf{X}))$. We select at most $M$ virtual cameras as in Eq.~\ref{eq:select}, with each point now weighted by its estimated reconstruction error:
\begin{equation}
    \mathcal{S} = \arg\max_{\substack{
        \mathcal{S}'\subseteq\mathcal{V}\\
        |\mathcal{S}'|\le M
    }}
    \sum_{\mathbf{X}\in\mathcal{P}_e}
    w(\mathbf{X}) \max_{v\in\mathcal{S}'} q_v(\mathbf{X}),
    \label{eq:select_virtual}
\end{equation}
which we solve with the same greedy selection.

\vspace{1mm}\noindent \textbf{Depth from Real--Virtual Pairs.} In contrast to a real--real pair, we construct each stereo pair from one real capture and one virtual rendering. The real capture serves as the reference view, and the second view is rendered from the proxy at the selected virtual camera. We feed the pair to the same stereo model and filter its prediction as in Sec.~\ref{subsec:real-real} to estimate a depth map $D_i^{v}$ for the captured view $I_i$, and denote by $\mathcal{S}_i \subseteq \mathcal{S}$ the selected virtual cameras attached to $I_i$.

\subsection{Training Loss}
\label{subsec:train_loss}

A captured image $I_i$ has a set of stereo depth maps $\mathcal{D}_i = \{D_i^{r}\} \cup \{D_i^{v}\}_{v \in \mathcal{S}_i}$. These depth maps come from stereo pairs with different baselines, and each is more accurate on different surfaces. At each pixel $p$, we keep the depth with the highest multi-view consistency:
\begin{equation}
\hat{D}_i(p) = \operatorname*{arg\,max}_{z \,\in\, \mathcal{D}_i(p)} \;
Q_i\big(p;\, z,\, \mathcal{Z}_i\big),
\label{eq:final_depth}
\end{equation}
where $\mathcal{D}_i(p)$ denotes the depths in $\mathcal{D}_i$ that are valid at pixel $p$. Here, $Q_i$ is computed as in Eq.~\ref{eq:vote}, but against the stereo depth maps $\mathcal{Z}_i = \bigcup_{k \in \mathcal{N}_i} \mathcal{D}_k$ of the neighboring views, from both real--real and real--virtual pairs, with the sum taken over all of these depth maps.

We compute the stereo loss between the rendered depth $\tilde{D}_i$ and
$\hat{D}_i$ in inverse depth space over the pixels $\Omega_i$ with valid stereo depth,
\begin{equation}
    \mathcal{L}_{\mathrm{stereo}}
    =
    \frac{1}{|\Omega_i|}
    \sum_{p \in \Omega_i}
    \left|
        \tilde{D}_i(p)^{-1}
        -
        \hat{D}_i(p)^{-1}
    \right|.
\end{equation}
Besides the stereo loss, we keep the normal consistency loss $\mathcal{L}_{n}$
from~\cite{2dgs}, and adopt from~\cite{metrogs} the appearance loss
$\mathcal{L}_{\mathrm{app}}$ for better appearance modeling and the scale
regularization $\mathcal{L}_{s}$ to limit oversized 2D Gaussians in large-scale
scenes. The final loss is
\begin{equation}
    \mathcal{L}
    =
    \mathcal{L}_{\mathrm{app}}
    + \lambda_{n} \mathcal{L}_{n}
    + \lambda_{s} \mathcal{L}_{s}
    + \lambda_{d} \mathcal{L}_{\mathrm{stereo}}.
    \label{eq:total}
\end{equation}

%% file: sec/5_experiments.tex
\section{Experiments}
\label{sec:exp}

\begin{table*}[t!]
    \centering
    \caption{Quantitative comparison of surface reconstruction
    quality on large-scale scenes. We report precision (P),
    recall (R), and F1 score. Higher is better.
    The best results are highlighted in bold red,
    while the second-best results are blue underlined.}
    \label{tab:gauuscene}
    \begin{tabular}{l|ccc|ccc|ccc}
        \toprule
        \multirow{2}{*}{Methods}
        & \multicolumn{3}{c|}{Russian Building}
        & \multicolumn{3}{c|}{Modern Building}
        & \multicolumn{3}{c}{Residence} \\
        \cmidrule(lr){2-4}
        \cmidrule(lr){5-7}
        \cmidrule(lr){8-10}
        & P$\uparrow$ & R$\uparrow$ & F1$\uparrow$
        & P$\uparrow$ & R$\uparrow$ & F1$\uparrow$
        & P$\uparrow$ & R$\uparrow$ & F1$\uparrow$ \\
        \midrule

        SuGaR \cite{sugar}
        & 0.480 & 0.369 & 0.417
        & 0.650 & 0.220 & 0.329
        & \second{0.579} & 0.287 & 0.384 \\

        GOF \cite{gof}
        & 0.294 & 0.394 & 0.330
        & 0.411 & 0.357 & 0.382
        & 0.404 & 0.418 & 0.411 \\

        2DGS \cite{2dgs}
        & 0.544 & 0.519 & 0.531
        & 0.588 & 0.413 & 0.485
        & 0.526 & 0.406 & 0.458 \\

        CityGaussian \cite{citygaussian}
        & 0.459 & 0.443 & 0.451
        & 0.582 & 0.381 & 0.461
        & 0.524 & 0.391 & 0.448 \\

        CityGS-X \cite{citygsx}
        & 0.572 & 0.516 & 0.542
        & 0.653 & 0.389 & 0.487
        & 0.512 & 0.411 & 0.456 \\

        CityGaussianV2 \cite{citygaussianv2}
        & 0.560 & 0.530 & 0.544
        & 0.643 & 0.398 & 0.492
        & 0.524 & 0.421 & 0.467 \\

        UrbanGS \cite{urbangs}
        & 0.568 & 0.525 & 0.546
        & 0.662 & 0.408 & 0.503
        & 0.529 & \best{0.461} & 0.493 \\

        MetroGS \cite{metrogs}
        & \second{0.591} & \second{0.571} & \second{0.580}
        & \second{0.667} & \second{0.429} & \second{0.522}
        & 0.566 & 0.439 & \second{0.494} \\

        \midrule
        \textbf{Ours}
        & \best{0.617} & \best{0.581} & \best{0.598}
        & \best{0.667} & \best{0.441} & \best{0.531}
        & \best{0.580} & \second{0.441} & \best{0.501}  \\

        \bottomrule
    \end{tabular}%
\end{table*}

\begin{table}[t!]
    \centering
    \caption{Comparison on the MatrixCity dataset.
    We report precision (P), recall (R), and F1 score.}
    \label{tab:matrixcity_geometry}
    \setlength{\tabcolsep}{4pt}
    \renewcommand{\arraystretch}{1.05}
    \resizebox{\columnwidth}{!}{%
    \begin{tabular}{l|ccc|ccc}
        \toprule
        \multirow{2}{*}{Methods}
        & \multicolumn{3}{c|}{MatrixCity-Aerial}
        & \multicolumn{3}{c}{MatrixCity-Street} \\
        \cmidrule(lr){2-4} \cmidrule(l){5-7}
        & P$\uparrow$ & R$\uparrow$ & F1$\uparrow$
        & P$\uparrow$ & R$\uparrow$ & F1$\uparrow$ \\
        \midrule
        SuGaR \cite{sugar}             & 0.182 & 0.157 & 0.169 & 0.053 & 0.111 & 0.071 \\
        GOF \cite{gof}                 & FAIL  & FAIL  & FAIL  & 0.219 & 0.473 & 0.300 \\
        2DGS \cite{2dgs}               & 0.207 & 0.390 & 0.270 & 0.334 & 0.659 & 0.443 \\
        CityGaussian \cite{citygaussian}     & 0.362 & 0.637 & 0.462 & 0.283 & 0.689 & 0.401 \\
        CityGS-X \cite{citygsx}   & 0.444 & \second{0.840} & 0.581 & OOM & OOM & OOM \\
        CityGaussianV2 \cite{citygaussianv2} & 0.441 & 0.752 & 0.556 & 0.376 & 0.759 & 0.503 \\
        MetroGS \cite{metrogs}         & \second{0.572} & 0.828 & \second{0.677}
                                       & \second{0.551} & \second{0.829} & \second{0.662} \\
        \midrule
        \textbf{Ours} & \best{0.620} & \best{0.865} & \best{0.723}
                      & \best{0.573} & \best{0.842} & \best{0.681} \\
        \bottomrule
    \end{tabular}}
    \vspace{-4mm}
\end{table}

\begin{figure*}[t!]
  \centering
  \includegraphics[width=\textwidth]{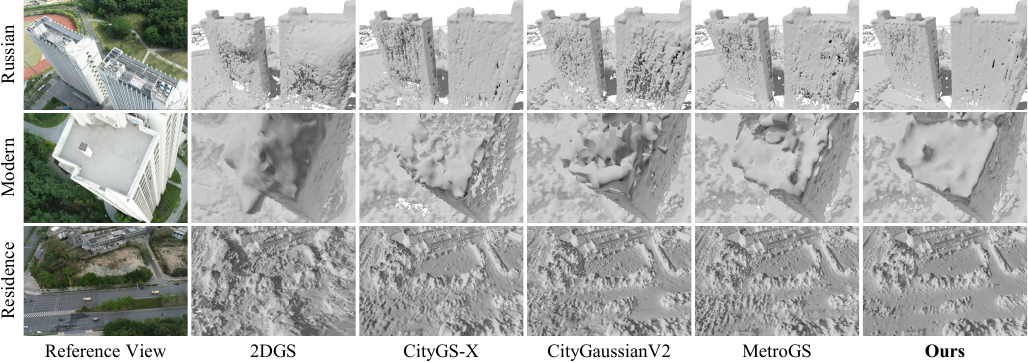}
  \caption{\textbf{Qualitative comparison of surface reconstruction.} We show shaded mesh reconstructions of Russian Building, Modern Building, and Residence from the GauU-Scene~\cite{gauuscene} dataset. Our method reconstructs more complete building facades, flatter roofs, and clearer and smoother street structures, respectively.}
  \label{fig:gauu}
\end{figure*}

\begin{figure*}[t!]
  \centering
  \includegraphics[width=\textwidth]{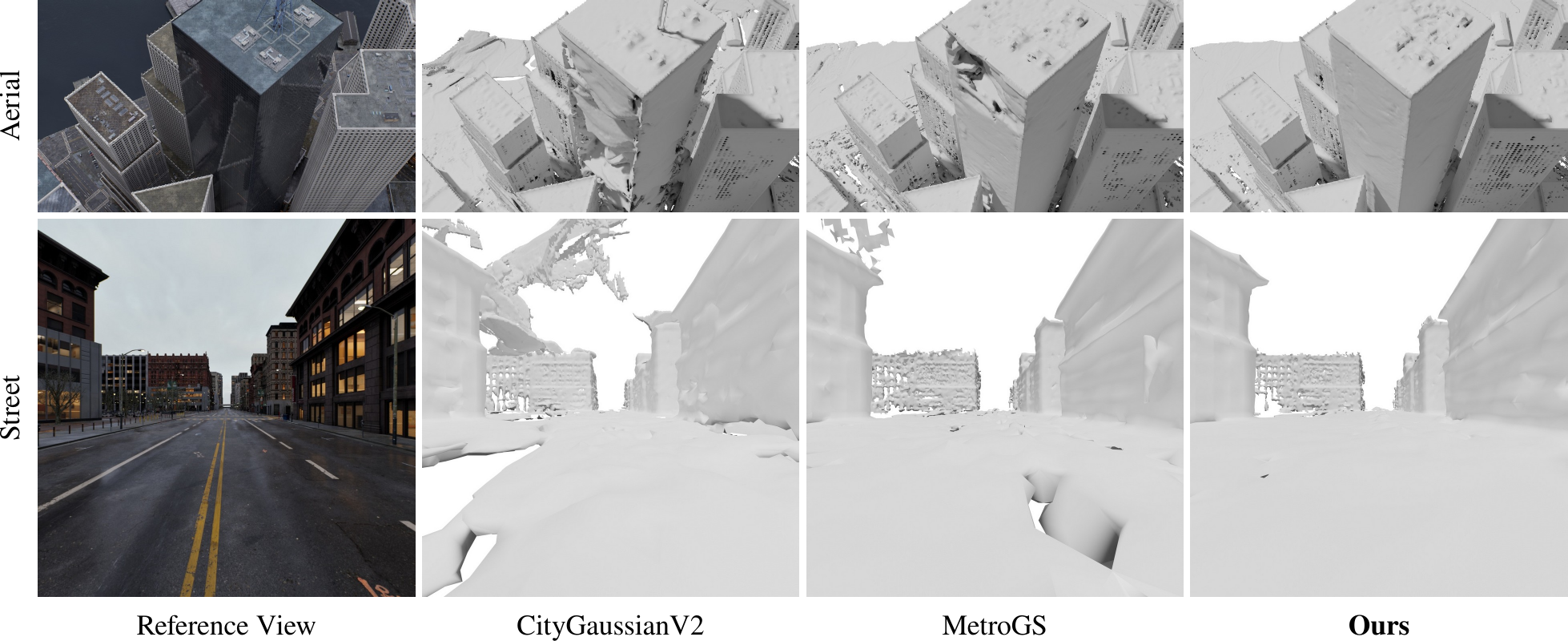}
  \caption{\textbf{Qualitative comparison of surface reconstruction on MatrixCity~\cite{matrixcity}.} It provides both aerial and street-level captures. In Aerial, our method reconstructs more complete facades of tall buildings. In Street, it recovers a more complete and flatter road surface.}
  \label{fig:matrix}
  \vspace{-4mm}
\end{figure*}

\noindent\textbf{Datasets and Metrics.} We evaluate TwinScape-GS on five scenes from the real-world GauU-Scene dataset \cite{gauuscene} and the synthetic MatrixCity dataset \cite{matrixcity}, following CityGaussianV2 \cite{citygaussianv2}. Both datasets have ground-truth point clouds, with LiDAR-based references for GauU-Scene and synthetic references for MatrixCity. To assess surface reconstruction quality, we extract meshes using TSDF from rendered median depth maps and report precision, recall, and F1 score.

\vspace{1mm}\noindent\textbf{Implementation Details.} We build our method on 2DGS~\cite{2dgs} and the MetroGS~\cite{metrogs} framework. Gaussians are initialized from a dense point cloud obtained by aligning a feed-forward pointmap (Pi3~\cite{pi3}) to the SfM points. For real stereo view pairing, we set $K=8$, the number of neighboring views commonly used in~\cite{pgsr,metrogs}, $d_{\min}=32$, and $d_{\max}=416$, the maximum disparity of FoundationStereo~\cite{foundationstereo}. For virtual stereo view planning, we sample two candidate virtual cameras for each captured view, with $\delta$ drawn uniformly from $[64, 128]$, and set the budget $M$ to the number of captured views in the scene. The loss weights $\lambda_n$ and $\lambda_s$ follow 2DGS and MetroGS, and $\lambda_d$ is 5 on GauU-Scene and 10 on MatrixCity. For each GauU-Scene~\cite{gauuscene} scene, we first train for 10K iterations without the stereo loss, then add the stereo depth from real--real pairs and train until 30K iterations. The model at 30K iterations serves as the proxy. We plan virtual cameras over it (Sec.~\ref{subsec:real-virtual}), add real--virtual pairs, and train until 60K iterations. On the much larger MatrixCity~\cite{matrixcity} dataset, we follow the same procedure but train with real--real pairs until 50K iterations, and train the Aerial and Street scenes for 150K and 180K iterations in total, respectively. All our models are trained with the distributed training of \cite{metrogs} on $4\times$ RTX 4090 GPUs with a batch size of 4.

\vspace{1mm}\noindent\textbf{Baselines.} We compare our method with state-of-the-art Gaussian-based reconstruction methods, including the large-scale approaches CityGaussian~\cite{citygaussian}, CityGaussianV2~\cite{citygaussianv2}, CityGS-X~\cite{citygsx}, UrbanGS~\cite{urbangs}, and MetroGS~\cite{metrogs}. We also compare against methods typically evaluated on smaller scenes, including SuGaR \cite{sugar}, GOF \cite{gof}, and 2DGS \cite{2dgs}.

\subsection{Comparisons}
\label{subsec:comparisons}

\noindent\textbf{Quantitative Results.} Tab.~\ref{tab:gauuscene} compares TwinScape-GS with state-of-the-art methods on the real-world GauU-Scene~\cite{gauuscene} dataset. Our method achieves the highest F1 score across all three scenes, demonstrating the best overall surface quality. Compared with MetroGS, TwinScape-GS improves the average F1 score from 0.532 to 0.543 and the average precision from 0.608 to 0.621, with the largest precision gain on Russian Building (0.591 to 0.617). On the larger-scale synthetic MatrixCity~\cite{matrixcity} dataset (Tab.~\ref{tab:matrixcity_geometry}), the improvements are even greater, and our method achieves the highest precision, recall, and F1 score on both aerial and street-level scenes. On MatrixCity-Aerial, our method improves the F1 score from 0.677 to 0.723 over MetroGS, with precision increasing from 0.572 to 0.620 and recall from 0.828 to 0.865. On MatrixCity-Street, the baselines of real--real pairs are dictated by the driving trajectory, and nearby surfaces often exceed the matching range of the stereo model. Our virtual stereo views are not limited by the trajectory, and the F1 score improves from 0.662 to 0.681.

\vspace{1mm}\noindent\textbf{Qualitative Results.} To further illustrate the geometric quality of our reconstructions, we provide visual comparisons in Fig.~\ref{fig:gauu} and Fig.~\ref{fig:matrix}. On GauU-Scene, existing methods produce pitted building facades, irregular rooftops, and noisy surfaces. In comparison, TwinScape-GS reconstructs flatter facades and rooftops with fewer geometric artifacts. On MatrixCity, CityGaussianV2 and MetroGS exhibit missing geometry and irregular surfaces on building facades in the aerial scene, as well as holes and distortions on the road in the street scene. TwinScape-GS produces more complete building facades and continuous road surfaces. Together with the quantitative results, these comparisons show that stereo depth from both paired captured views and planned virtual cameras improves the accuracy and completeness of the reconstructed surfaces.

\subsection{Ablation Studies}

We evaluate our design choices on the Russian Building scene of GauU-Scene~\cite{gauuscene}, training every model for 60K iterations.

Tab.~\ref{tab:component_ablation} compares three configurations with the default settings. The baseline is our model trained without the stereo loss $\mathcal{L}_{\mathrm{stereo}}$ in Eq.~\ref{eq:total}. Supervising with the stereo depth from real--real pairs (Sec.~\ref{subsec:real-real}) alone raises F1 from 55.95 to 59.13, improving both precision and recall. The full method further adds real--virtual pairs (Sec.~\ref{subsec:real-virtual}) and raises precision from 60.23 to 61.67, reaching the best F1 of 59.81. The planned virtual views give reliable stereo depth on surfaces where real--real pairs do not, which makes the reconstructed surfaces more accurate.

\begin{table}[t!]
    \centering
    \caption{
        Ablation study of the proposed main components.
        We report precision (P), recall (R), and F1 score.
        Best results are shown in bold. All values are in percent.
    }
    \label{tab:component_ablation}

    \begin{tabular}{l|ccc}
        \toprule
        Configuration
        & P $\uparrow$
        & R $\uparrow$
        & F1 $\uparrow$ \\
        \midrule

        Baseline
        & 55.83 & 56.08 & 55.95  \\

        + Real--Real Pairs
        & 60.23 & \textbf{58.07} & 59.13   \\

        Full method (+Real--Virtual Pairs)
        & \textbf{61.67} & 58.06 & \textbf{59.81}  \\

        \bottomrule
    \end{tabular}
\end{table}

\vspace{1mm}\noindent \textbf{Neighboring View Selection.} We further validate our disparity-based stereo pairing (Sec.~\ref{subsec:real-real}) in Tab.~\ref{tab:ablation_pairing}. All models are trained with the stereo depth from real--real pairs only, and differ only in how the neighboring views are selected. The first row follows MetroGS~\cite{metrogs}, which selects the nearest covisible views. Selecting views by disparity raises recall from 56.99 to 57.87, as more surfaces get a large disparity. Adding the upper bound $d_{\max}$ gives the larger part of the remaining gain, since disparities beyond it cannot be matched by the stereo model, and the lower bound $d_{\min}$ brings a further small improvement. The full pair score raises F1 from 58.15 to 59.13.

\begin{table}[t]
\centering
\caption{Ablation on the neighboring view selection strategy (Sec.~\ref{subsec:real-real}). All models are trained with the stereo depth from real--real pairs. All values are in percent.}
\label{tab:ablation_pairing}
\begin{tabular}{l|ccc}
\toprule
Neighboring view selection & P $\uparrow$ & R $\uparrow$ & F1 $\uparrow$ \\
\midrule
Nearest covisible views          & 59.36    & 56.99    & 58.15    \\
Ours, w/o disparity range        & 59.54    & 57.87    & 58.70    \\
Ours, w/o $d_{\min}$             & 59.94    & 58.07    & 58.99    \\
Ours                             & \textbf{60.23} & \textbf{58.07} & \textbf{59.13} \\
\bottomrule
\end{tabular}
\end{table}

\vspace{1mm}\noindent \textbf{Virtual Camera Selection.} Tab.~\ref{tab:ablation_planning} tests two designs of the planner. The first is whether a captured view needs more than one candidate virtual camera. The second is whether the reconstruction error estimated from multi-view consistency is useful for the selection. By default, we oversample two candidate virtual cameras for each captured view and select 500 of them. With only one candidate per captured view, F1 falls from 59.81 to 59.40, as the planner can no longer choose between baselines. Without the estimated reconstruction error, it falls to 59.35. The estimated error is therefore what directs the virtual cameras to the surfaces that need them.

\begin{table}[t]
\centering
\caption{Ablation on virtual camera selection (Sec.~\ref{subsec:real-virtual}). ``One candidate per view'' samples a single candidate virtual camera for each captured view. ``W/o estimated error'' selects virtual cameras without the estimated reconstruction error. All values are in percent.}
\label{tab:ablation_planning}
\begin{tabular}{l|ccc}
\toprule
Virtual camera selection & P $\uparrow$ & R $\uparrow$ & F1 $\uparrow$ \\
\midrule
One candidate per view      & 61.24 & 57.67 & 59.40 \\
W/o estimated error         & 60.89 & 57.88 & 59.35 \\
Ours                        & \textbf{61.67} & \textbf{58.06} & \textbf{59.81} \\
\bottomrule
\end{tabular}
\end{table}

\begin{figure}[t]
  \centering
  \includegraphics[width=0.75\linewidth]{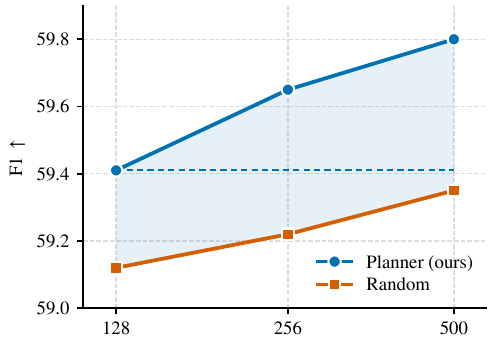}
  \caption{Planner vs.\ random selection under different budgets $M$ of virtual cameras.}
  \label{fig:ablation_budget}
  \vspace{-4mm}
\end{figure}

\vspace{1mm}\noindent \textbf{Planner vs. Random Selection.} To examine how well the planner selects virtual cameras, we compare it with random selection from the same candidates under different budgets in Fig.~\ref{fig:ablation_budget}.  With about a quarter of the largest budget, the planner already exceeds random selection with the largest budget. The planner is better under every budget, and the gap widens as the budget grows. The largest budget equals the number of captured views in this scene. Since a large-scale scene can only afford a limited number of virtual cameras, the planner makes better use of the same budget.

%% file: sec/6_conclusion.tex
\section{Conclusion}
\label{sec:conclusion}
In this paper, we present TwinScape-GS, a 2DGS-based method for city-scale surface reconstruction through virtual flyovers. Building on a pretrained stereo model, we introduce a disparity-based stereo pairing scheme that builds a proxy from the captured views, and a multi-view consistency-guided virtual flyover planner that allocates virtual cameras to poorly reconstructed surfaces. Together, they recover poorly constrained surfaces from new viewing configurations. Experiments on large-scale urban scenes validate our method, showing accurate and complete surface reconstruction.

\vspace{1mm}\noindent\textbf{Limitations and future work.} Like other methods that render virtual views from a Gaussian model, TwinScape-GS cannot place virtual cameras too far from the captured views, because renderings there may be degraded for stereo matching. Nevertheless, our work  also offers a new perspective for city-scale capture. In practice, virtual views might be planned on the reconstruction to complement the physical capture, without the cost of a second flight.

%% file: sec/7_supplementary.tex
\clearpage
\onecolumn

\begin{center}
    {\Large\bfseries Supplementary Material\par}
\end{center}

\vspace{1em}
\appendix

\section{More Implementation Details}

\noindent {\bf Data Preprocessing and Splits.}
We follow CityGaussianV2~\cite{citygaussianv2} for data preprocessing and data splits. For GauU-Scene~\cite{gauuscene}, we use the camera poses and SfM points provided by the dataset. We downsample the images by a factor of 3.4175, so that the longer side is 1600 pixels. We then sort the images in COLMAP order and hold out every ninth image for testing. For MatrixCity~\cite{matrixcity}, we use the training and test split provided by the dataset. We downsample the aerial images to $1600\times900$ and keep the street images at their original $1000\times1000$ resolution. For the ground truth, GauU-Scene provides LiDAR point clouds aligned to the COLMAP coordinate system, and MatrixCity provides point clouds of the synthetic scenes. We use the ground-truth point clouds processed by CityGaussianV2. They are downsampled and cropped to the region covered by the training views.

\vspace{1mm}\noindent {\bf Stereo Matching.}
We use the same pretrained FoundationStereo~\cite{foundationstereo} model for both real--real and real--virtual pairs. The stereo model runs at the training image resolution, and $d_{\min}$ and $d_{\max}$ are given in pixels at this resolution. A virtual camera $v$ keeps the orientation and intrinsics of its captured view $I_i$, and its center is
\begin{equation}
    \mathbf{c}_v = \mathbf{c}_i + b_v\,\mathbf{R}_i\,[1, 0, 0]^{\top},
\end{equation}
where $\mathbf{c}_i$ and $\mathbf{R}_i$ are the camera center and the camera-to-world rotation of $I_i$, and $b_v$ is the baseline of $v$. A real--virtual pair therefore needs no rectification. For real--real pairs, we map the depth from the rectified view back to the reference view.

\vspace{1mm}\noindent {\bf Virtual Stereo View Planning.}
The planning runs once offline on the proxy at 30K iterations (50K on MatrixCity). In Eq.~\ref{eq:vote}, a pixel $p$ has a valid projection in a neighboring view $I_k$ if its warped point lies in front of the camera of $I_k$ and inside the image, and all pixels used for bilinear interpolation have a valid depth. For efficiency, we build the surface points $\mathcal{P}_e$ from every second pixel, voxelize them with a voxel size of 0.01 scene units, keep the largest weight $w$ within each voxel, and evaluate Eq.~\ref{eq:select_virtual} on the voxels.

\vspace{1mm}\noindent {\bf MatrixCity-Street.}
Unlike the aerial scenes, MatrixCity-Street is captured by cameras that move forward along the roads. The baselines of real--real pairs therefore follow the driving direction, and nearby surfaces such as the road often exceed the matching range of the stereo model (Sec.~\ref{subsec:comparisons}). On this scene, we train with real--real pairs until 50K iterations as on MatrixCity-Aerial, and then supervise the rest of the training with real--virtual pairs alone. We keep all candidate virtual cameras, with up to two baselines for each captured view. The short baseline is set from the nearest depths in the view, so that the road close to the camera has a disparity below $d_{\max}$. The long baseline is set from the median depth of the view and gives distant surfaces a larger disparity.

\vspace{1mm}\noindent {\bf Hyperparameters.}
We set $\lambda_n = 0.0125$ and $\lambda_s = 0.1$ on all scenes. Following MetroGS~\cite{metrogs}, we keep its default setting for the appearance loss and stop densification at iteration 15K on GauU-Scene and 50K on MatrixCity. Following CityGaussianV2~\cite{citygaussianv2}, TSDF fusion uses a voxel size of 0.01 and a truncation distance of 0.04 in scene units on GauU-Scene, and ignores depths beyond 2.0. MatrixCity-Aerial uses the same voxel size and truncation distance and ignores depths beyond 5.0. On MatrixCity-Street, the voxel size, truncation distance, and depth limit are 1, 4, and 500, respectively.

\section{Evaluation Metrics}

We follow the evaluation protocol of CityGaussianV2~\cite{citygaussianv2} and report precision, recall, and F1 score~\cite{knapitsch2017tanks} as fractions in $[0,1]$ in the comparison tables of the main paper.

 We extract the mesh by TSDF fusion of the depth maps rendered at the training views. We then uniformly sample points on the mesh surface. These points and the ground-truth point cloud are cropped and downsampled before computing distances. We denote the resulting point sets by $\mathcal{X}_{\mathrm{rec}}$ and $\mathcal{X}_{\mathrm{gt}}$, respectively. The distance from a point $\mathbf{x}$ to a point set $\mathcal{X}$ is
\begin{equation}
    \mathrm{dist}(\mathbf{x}, \mathcal{X}) = \min_{\mathbf{y}\in\mathcal{X}} \|\mathbf{x}-\mathbf{y}\|_2 .
\end{equation}
Precision is the fraction of reconstructed points whose distance to the ground truth is below a threshold $\tau$:
\begin{equation}
    P = \frac{1}{|\mathcal{X}_{\mathrm{rec}}|} \sum_{\mathbf{x}\in\mathcal{X}_{\mathrm{rec}}} \mathbf{1}\!\left[\mathrm{dist}(\mathbf{x}, \mathcal{X}_{\mathrm{gt}}) < \tau\right].
\end{equation}
Recall is the fraction of ground-truth points whose distance to the reconstruction is below $\tau$:
\begin{equation}
    R = \frac{1}{|\mathcal{X}_{\mathrm{gt}}|} \sum_{\mathbf{x}\in\mathcal{X}_{\mathrm{gt}}} \mathbf{1}\!\left[\mathrm{dist}(\mathbf{x}, \mathcal{X}_{\mathrm{rec}}) < \tau\right].
\end{equation}
The F1 score is the harmonic mean of precision and recall:
\begin{equation}
    F_1 = \frac{2PR}{P+R}.
\end{equation}
The threshold $\tau$ is set for each scene as in~\cite{citygaussianv2}.

\section{More Quantitative Results}

\noindent \textbf{Chamfer Distance.} We report the Chamfer-$L_1$~\cite{peng2020convolutional} and Chamfer-$L_2$ distances on GauU-Scene and MatrixCity-Aerial, computed on the same point sets $\mathcal{X}_{\mathrm{rec}}$ and $\mathcal{X}_{\mathrm{gt}}$. For $k \in \{1,2\}$, we compute
\begin{equation}
    \begin{aligned}
        \mathrm{CD}_{L_k} = \frac{1}{2}\Bigg(
        &\frac{1}{|\mathcal{X}_{\mathrm{rec}}|}
        \sum_{\mathbf{x}\in\mathcal{X}_{\mathrm{rec}}}
        \mathrm{dist}(\mathbf{x}, \mathcal{X}_{\mathrm{gt}})^k \\
        +{}&\frac{1}{|\mathcal{X}_{\mathrm{gt}}|}
        \sum_{\mathbf{x}\in\mathcal{X}_{\mathrm{gt}}}
        \mathrm{dist}(\mathbf{x}, \mathcal{X}_{\mathrm{rec}})^k
        \Bigg).
    \end{aligned}
\end{equation}
As shown in Tab.~\ref{tab:supp_chamfer}, TwinScape-GS achieves the lowest Chamfer-$L_1$ and Chamfer-$L_2$ distances on all three scenes of GauU-Scene. On MatrixCity-Aerial, it achieves the lowest Chamfer-$L_1$ distance and the second-lowest Chamfer-$L_2$ distance.

\begin{table*}[ht]
    \centering
     \caption{Chamfer distance on GauU-Scene and MatrixCity-Aerial.
    We report the Chamfer-$L_1$ (CD-$L_1$) and Chamfer-$L_2$ (CD-$L_2$)
    distances in the coordinates used for evaluation, both multiplied by $10^3$.
    The best results are highlighted in bold red,
    and the second-best results are blue underlined.}
    \label{tab:supp_chamfer}
    \begin{tabular}{l|cc|cc|cc|cc}
        \toprule
        \multirow{2}{*}{Methods}
        & \multicolumn{2}{c|}{Russian Building}
        & \multicolumn{2}{c|}{Modern Building}
        & \multicolumn{2}{c|}{Residence}
        & \multicolumn{2}{c}{MatrixCity-Aerial} \\
        \cmidrule(lr){2-3}
        \cmidrule(lr){4-5}
        \cmidrule(lr){6-7}
        \cmidrule(lr){8-9}
        & CD-$L_1$$\downarrow$ & CD-$L_2$$\downarrow$
        & CD-$L_1$$\downarrow$ & CD-$L_2$$\downarrow$
        & CD-$L_1$$\downarrow$ & CD-$L_2$$\downarrow$
        & CD-$L_1$$\downarrow$ & CD-$L_2$$\downarrow$ \\
        \midrule

        CityGaussianV2 \cite{citygaussianv2}
        & 7.901 & 0.504
        & 11.457 & 1.290
        & 6.835 & 0.210
        & 12.061 & \best{0.556} \\

        MetroGS \cite{metrogs}
        & \second{7.060} & \second{0.420}
        & \second{6.209} & \second{0.122}
        & \second{5.874} & \second{0.142}
        & \second{8.543} & 0.655 \\

        \midrule
        \textbf{Ours}
        & \best{6.713} & \best{0.377}
        & \best{6.058} & \best{0.114}
        & \best{5.614} & \best{0.124}
        & \best{8.271} & \second{0.649} \\

        \bottomrule
    \end{tabular}
\end{table*}

\noindent \textbf{Rendering Quality.} We report the rendering results in Tabs.~\ref{tab:gauuscene_rendering} and~\ref{tab:matrixcity_rendering}. Although our method focuses on geometric reconstruction rather than rendering, its rendering quality is competitive with state-of-the-art city-scale methods. Since we adopt the appearance loss of MetroGS~\cite{metrogs}, our results are close to those of MetroGS.

\begin{table}[h]
    \centering
    \caption{Quantitative comparison of rendering quality
    on GauU-Scene. We compare with city-scale methods
    and 2DGS, on which our method is built.
    We report PSNR, SSIM, and LPIPS.
    Higher PSNR and SSIM and lower LPIPS are better.
    The best results are highlighted in bold red,
    and the second-best results are blue underlined.}
    \label{tab:gauuscene_rendering}
    \begin{tabular}{l|ccc|ccc|ccc}
        \toprule
        \multirow{2}{*}{Methods}
        & \multicolumn{3}{c|}{Russian Building}
        & \multicolumn{3}{c|}{Modern Building}
        & \multicolumn{3}{c}{Residence} \\
        \cmidrule(lr){2-4}
        \cmidrule(lr){5-7}
        \cmidrule(lr){8-10}
        & PSNR$\uparrow$ & SSIM$\uparrow$ & LPIPS$\downarrow$
        & PSNR$\uparrow$ & SSIM$\uparrow$ & LPIPS$\downarrow$
        & PSNR$\uparrow$ & SSIM$\uparrow$ & LPIPS$\downarrow$ \\
        \midrule

        2DGS \cite{2dgs}
        & 23.77 & 0.788 & 0.189
        & 25.77 & 0.776 & 0.202
        & 22.24 & 0.703 & 0.306 \\

        CityGaussian \cite{citygaussian}
        & 24.37 & 0.808 & 0.163
        & 26.29 & 0.796 & 0.160
        & 23.59 & 0.763 & 0.204 \\

        CityGS-X \cite{citygsx}
        & 24.62 & 0.804 & 0.155
        & 26.20 & 0.783 & 0.171
        & 23.74 & 0.749 & 0.238 \\

        CityGaussianV2 \cite{citygaussianv2}
        & 24.12 & 0.784 & 0.196
        & 25.84 & 0.770 & 0.207
        & 23.57 & 0.742 & 0.243 \\

        UrbanGS \cite{urbangs}
        & 24.53 & 0.810 & 0.158
        & 26.44 & \best{0.805} & 0.157
        & 23.78 & 0.762 & 0.206 \\

        MetroGS \cite{metrogs}
        & \second{24.94} & \second{0.814} & \best{0.138}
        & \best{27.07} & 0.797 & \best{0.152}
        & \second{24.51} & \second{0.769} & \second{0.185} \\

        \midrule
        \textbf{Ours}
        & \best{25.05} & \best{0.818} & \second{0.140}
        & \second{26.79} & \second{0.801} & \second{0.155}
        & \best{24.51} & \best{0.779} & \best{0.185} \\

        \bottomrule
    \end{tabular}
\end{table}

\begin{table}[t]
    \centering
    \caption{Quantitative comparison of rendering quality
    on MatrixCity-Aerial. We compare with city-scale methods
    and 2DGS, on which our method is built.
    We report PSNR, SSIM, and LPIPS.
    Higher PSNR and SSIM and lower LPIPS are better.
    The best results are highlighted in bold red,
    and the second-best results are blue underlined.
    ``--'' means that the metric is not reported in the original paper.}
    \label{tab:matrixcity_rendering}
    \begin{tabular}{l|ccc}
        \toprule
        Methods
        & PSNR$\uparrow$
        & SSIM$\uparrow$
        & LPIPS$\downarrow$ \\
        \midrule

        2DGS \cite{2dgs}
        & 21.35 & 0.632 & 0.562 \\

        CityGaussian \cite{citygaussian}
        & 27.46 & \best{0.865} & 0.204 \\

        CityGS-X \cite{citygsx}
        & \best{27.58} & -- & -- \\

        CityGaussianV2 \cite{citygaussianv2}
        & 27.23 & \second{0.857} & \second{0.169} \\

        MetroGS \cite{metrogs}
        & 27.52 & 0.854 & \best{0.167} \\

        \midrule
        \textbf{Ours}
        & \second{27.53} & 0.852 & 0.172 \\

        \bottomrule
    \end{tabular}
\end{table}

\section{More Qualitative Results}

\begin{figure*}[t!]
  \centering
  \includegraphics[width=\textwidth]{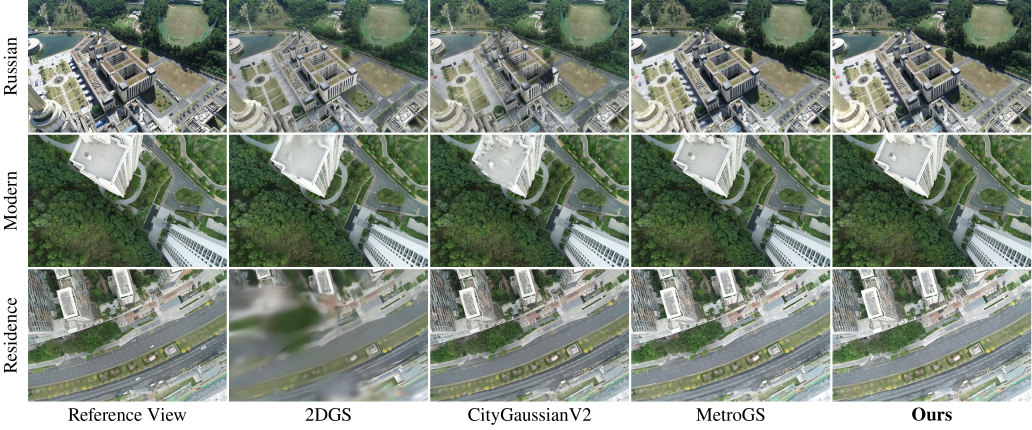}
    \caption{\textbf{Qualitative comparison of rendering.} We show images rendered at test views of Russian Building, Modern Building, and Residence from the GauU-Scene~\cite{gauuscene} dataset.}
  \label{fig:rgb_sup}
\end{figure*}

Fig.~\ref{fig:rgb_sup} compares the images rendered at test views by our method with those of 2DGS~\cite{2dgs}, CityGaussianV2~\cite{citygaussianv2}, and MetroGS~\cite{metrogs} on GauU-Scene~\cite{gauuscene}.

